\documentclass[conference]{IEEEtran}

\usepackage{balance}

\IEEEoverridecommandlockouts 
\usepackage[dvips]{graphicx}
\usepackage{algorithmic}
\usepackage{algorithm}
\usepackage{listings}
\usepackage[OT4,T1]{fontenc}
\usepackage[cmex10]{amsmath}
\usepackage{url}
\usepackage{multirow}

\usepackage{todonotes}
\usepackage{amssymb}
\usepackage{url}
\usepackage{amsmath}

\title{Formal analysis of HTM Spatial Pooler performance under predefined operation conditions}
\author{
\IEEEauthorblockN{Marcin Pietro\'n, Maciej Wielgosz, Kazimierz Wiatr}
\IEEEauthorblockA{
AGH University of Science and Technology\\
al. Mickiewicza 30, 30-059 Krakow, Poland\\
Email: {pietron,wielgosz,wiatr}@agh.edu.pl}

}

\begin{document}
\maketitle              

\begin{abstract}
This paper introduces mathematical formalism for Spatial (SP) of Hierarchical Temporal Memory (HTM) with a spacial consideration for its hardware implementation. Performance of HTM network and its ability to learn and adjust to a problem at hand is governed by a large set of parameters. Most of parameters are codependent which makes creating efficient HTM-based solutions challenging. It requires profound knowledge of the settings and their impact on the performance of system. Consequently, this paper introduced a set of formulas which are to facilitate the design process by enhancing tedious trial-and-error method with a tool for choosing initial parameters which enable quick learning convergence. This is especially important in hardware implementations which are constrained by the limited resources of a platform.

The authors focused especially on a formalism of Spatial Pooler and derive at the formulas for quality and convergence of the model. This may be considered as recipes for designing efficient HTM models for given input patterns.

\end{abstract}

\section{Introduction}
\label{section:introduction}
Recent years witnessed huge progress in deep learning architecture driven mostly by abundance of training data and huge performance of parallel GPU processing units \cite{Liu}\cite{Bengio}. This sets a new path in a development of intelligent systems and ultimately puts us on a track to general artificial intelligence solutions. It is worth noting that in addition to well-established Convolutional Neural Networks (CNN) architectures there is a set of biologically inspired solutions such as Hierarchical Temporal Memories \cite{Ahmad}\cite{How_neurons}\cite{Chen}. Those architectures as well as CNNs suffer from lack of well-defined mathematical formulation of rules for their efficient hardware implementation. Large range of heuristics and rules of thumb are used instead. This was not very harmful except for a long training time when most of the algorithm were executed on CPUs without hardware acceleration. However, nowadays most of biologically inspired are ported to hardware for a sake of performance efficiency \cite{Woodbeck}\cite{Thomas}. This in turn requires a profound consideration and analysis of resources consumption to be able to predict both the capacity of the platform and the ultimate performance of the system. Consequently, the authors of the papers analyzed HTM design constrains on the mathematical ground and formulated a set of directives for building hardware modules. 

The rest of the paper is organized as follows. Section \ref{section:htm} provides the background and related works. Section \ref{section:mathematical_formalism} describes mathematical formalism of Spatial Pooler. Finally, we present our conclusions in Section \ref{section:conclusions}.

\section{HTM architecture}
\label{section:htm}
Hierarchical Temporal Memory (HTM) replicates the structural and algorithmic properties of the neocortex \cite{Mountcastle}. It can be regarded as a memory system which is not programmed and it is trained through exposing them to data i.e. text. HTM is organized in the hierarchy which reflects the nature of the world and performs modeling by updating the hierarchy. The structure is hierarchical in both space and time, which is the key in natural language modeling since words and sentences come in sequences which describe cause and effect relationships between the latent objects.  HTMs may be considered similar to Bayesian Networks, HMM and Recurrent Neural Networks, but they are different in the way hierarchy, model of neuron and time is organized.

At any moment in time, based on current and past input, an HTM will assign a likelihood that given concepts are present in the examined stream. The HTM's output constitutes a set of probabilities for each of the learned causes. This moment-to-moment distribution of possible concepts (causes) is denoted as a belief. If the HTM covers a certain number of concepts it will have the same number of variables representing those concepts. Typically HTMs learn about many causes and create a structure of them which reflects their relationships.

Even for human beings, discovering causes is considered to be a core of perception and creativity, and people through course of their life learn how to find causes underlying objects in the world. In this sense HTMs mimic human cognitive abilities and with a long enough training, proper design and implementation, they should be able to discover causes humans can find difficult or are unable to detect.

HTM infers concepts of new stream elements and the result is a distribution of beliefs across all the learned causes. If the concept (e.g. one of the categories occurring in the examined stream) is unambiguous, the belief distribution will be peaked otherwise it will be flat. In HTMs it is possible to disable learning after training and still do inference.

\subsection{Encoder}
The role of an encoder within HTM processing flow is critical. It maps data from various representations to Sparse Distributed Representation (SDR) which is an internal form of holding data within HTM network. Quality of a transformation process directly affect the performance of the implemented system. The conversion involves mapping to strongly semantic-oriented way of representing data in which a single bit holds a meaning of tens or hundreds of original representation bits. 
There are various kinds of encoders for different input data \cite{Purdy}.

\subsection{Spatial Pooler}

\begin{figure*}[tbp]
\centering
\includegraphics[width=1\hsize]{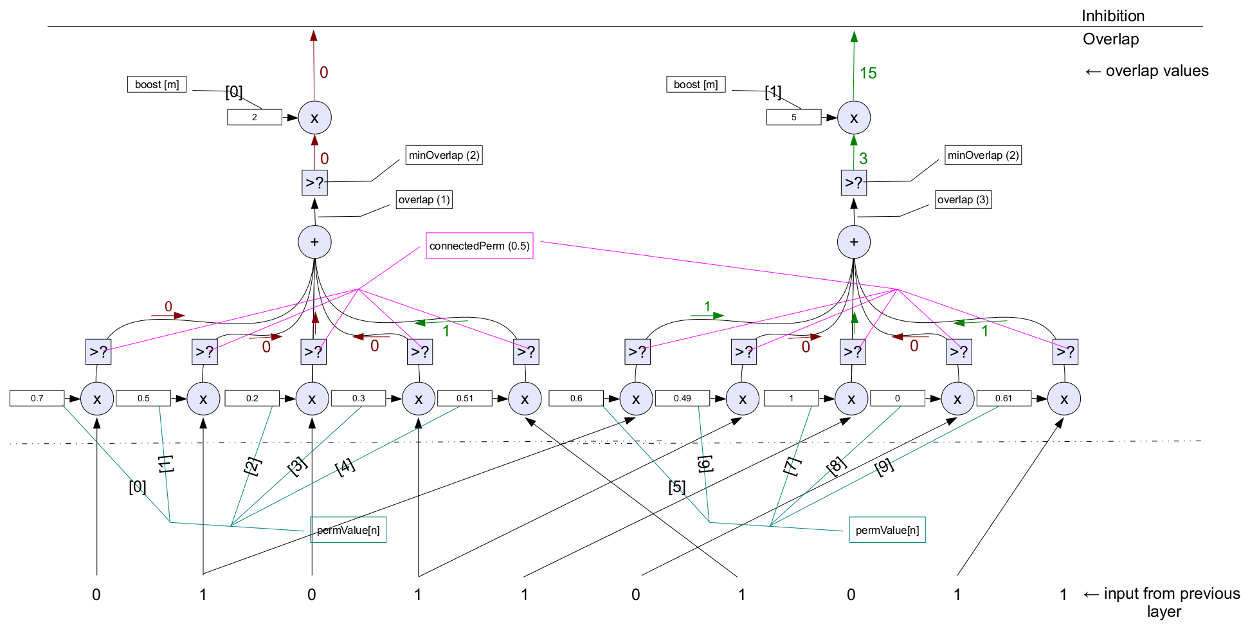}
\caption{Architecture of HTM Spatial Pooler}
\label{fig:SP_architecure}
\end{figure*}

SP operates in the spatial domain and acts as an advanced encoder which translates from binary representation to the sparse distributed binary representation of approx. 3\% density. SP is also an interface between the encoder and the remaining part of the HTM module Fig. \ref{fig:SP_architecure}.

The detailed implementation of the algorithm is as follows:

\begin{itemize}
 \item Each column is connected by a fixed number of inputs to randomly selected node inputs. Based on the input pattern, some columns will receive more active input values,
 \item Inputs of columns (synapses) have values (floating point between 0 and 1 called permanence value) which represents possibility of activating the synapse (if the value is greater than 0.5 and corresponding input is 1 the synapse is active),
 \item Columns which have more active connected synapses than given threshold (minOverlap) and its overlap parameter (number of active synapses) is better than k-th overlap of set of columns in spectrum of inhibition radius,
 \item During learning process columns gather information about their history of activity, overlap values (if overlap was greater than minOverlap or not), compute minimum and maximum overlap duty cycle and then decides according to the combinations of these parameters if their permanence values or inhibition radius should be changed.
\end{itemize}

There are two main parameters of Spatial Pooler which mimic the behaviour of the mammalian brain, i.e. permValue and inhibitionRadius. The first one reflects the sensitivity of the brain to external stimuli, and the latter one may be considered a focus. In the process of adjusting those parameters, SP adapts to the profile of the input data.

Generally, spatial pooling selects a relatively constant number of the most active columns and inactivates (inhibits) other columns in the vicinity of the active ones. Similar input patterns tend to activate a stable set of columns. The classifier module based on Spatial Pooler is realized by overlap and activation computing on incoming input values. The rest of the parameters are set during learning process. The functionality of Spatial Pooler is similar to LVQ or Self Organizing Maps neural models.

\begin{algorithm}[tbp]
\label{alg:1}
\caption{Overlap }
\begin{algorithmic}[2]
\FORALL{col $\in$ sp.columns} 
\STATE{col.overlap $\leftarrow$ 0} 

\FORALL{syn $\in$ col.connected\_synapses()} 
\STATE{col.overlap $\leftarrow$ col.overlap + syn.active()} 
\ENDFOR

\IF{col.overlap < min\_overlap}
\STATE col.overlap $\leftarrow$ 0;
\ELSE
\STATE col.overlap $\leftarrow$ col.overlap * col.boost;
\ENDIF

\ENDFOR
\end{algorithmic}
\end{algorithm}

\begin{algorithm}[tbp]
\label{alg:2}
\caption{Inhibition}
\begin{algorithmic}[2]
\FORALL{col $\in$ sp.columns} 
\STATE{max\_column $\leftarrow$ max(n\_max\_overlap(col, n), 1)} 

\IF{col.overlap > max\_column}
\STATE col.active $\leftarrow$ 1;
\ELSE
\STATE col.active $\leftarrow$ 0;
\ENDIF

\ENDFOR
\end{algorithmic}
\end{algorithm}

\begin{algorithm}[tbp]
\label{alg:3}
\caption{Learning : adapting perm values}
\begin{algorithmic}[2]
\FORALL{col $\in$ sp.columns} 
\STATE{max\_column $\leftarrow$ max(n\_max\_overlap(col, n), 1)} 

  \IF{col.active}
    \FORALL{synapses $\in$ col.synapses} 
      \IF { syn.active()}
	\STATE syn.perm\_value $\leftarrow$ min(1, syn.perm\_value + syn.perm\_inc);
	\ELSE
	\STATE syn.perm\_value $\leftarrow$ max(0, syn.perm\_value - syn.perm\_dec);
      \ENDIF
    \ENDFOR
  \ENDIF
\ENDFOR
\end{algorithmic}
\end{algorithm}

\begin{algorithm}[tbp]
\label{alg:4}
\caption{Learning : Column boosting operation}
\begin{algorithmic}[2]
\FORALL{col $\in$ sp.columns} 
\STATE{col.mdc $\leftarrow$ 0.01 * max\_adc(range\_neighbor(col))}
\STATE{col.update\_active\_duty\_cycle()} 
\STATE{col.update\_boost()}
\STATE{col.update\_overlap\_duty\_cycle()
} 
  \IF{col.odc < col.mdc}
    \FORALL{synapses $\in$ col.synapses} 
      \STATE syn.perm\_value $\leftarrow$ min(1, syn.perm\_value + 0.1 * min\_perm\_value);
    \ENDFOR
  \ENDIF
\ENDFOR

sp.update\_inhibition\_radius()
\end{algorithmic}
\end{algorithm}

\section{Mathematical formalism}
\label{section:mathematical_formalism}
\subsection{Spatial Distributed Representation}
This section covers properties and important features of SDR (Sparse Distributed Representation) vector space.

\subsection{Space definition}

\begin{equation}
S: \{0,1\}^n
\label{eq:sdr_space}
\end{equation}
where $S$ and $n$ are SDR vector space and its dimension respectively.

Depending on the context, the vectors are called points, patterns or words.

\subsection{Notation}
This section covers HTM encoder formalism. It is mostly based on \cite{Purdy}.
\begin{itemize}
 \item $A$ - an arbitrary input set,
 \item $a$ - element of the input space $A$,
 \item $S(n,k)$ - SDR (Sparse Distributed Representation) of length $n$ and $w$ bits on (egual $1$),
 \item $s$ - element of SDR space $S$,
 \item $n$ - total number of bits in a vector of SDR ($s$),
 \item $w$ - number of active bits in a vector of SDR ($s$),
 \item $buckets$ - number of buckets (ranges) to which $a_i$ is mapped $S$,
 \item $val_{min}$ - min value of the input space range,
 \item $val_{max}$ - max value of the input space range.
\end{itemize}

\subsection{Preserving semantics}
Both input space and SDR space are decent metric spaces which means that they meet metric space postulates \cite{Metri72:online}.

\begin{itemize}
 \item $(A,d_A)$ - input space
 \item $d_A$ - input space metric
 \item $(S,d_S)$ - SDR space
 \item $d_S$ - SDR space metric
\end{itemize}
 
\subsubsection{Input space}
 
\begin{equation}
d_A:A \times A \rightarrow \mathbb{R}
\label{eq:metric_space_postulate_function_A}
\end{equation}

\begin{equation}
\bigwedge_{a_i, a_j \in A} d_A(a_i, a_j)  \geqslant 0
\label{eq:metric_space_postulate_separation_A}
\end{equation}

\begin{equation}
\bigwedge_{a_i, a_j \in A} d_A(a_i, a_j) = 0 \Longleftrightarrow a_i = a_j
\label{eq:metric_space_identyity_A}
\end{equation}

\begin{equation}
\bigwedge_{a_i, a_j \in A} d_A(a_i, a_j) = d_A(a_j, a_i)
\label{eq:metric_space_symmetry_A}
\end{equation}

\begin{equation}
\bigwedge_{a_i, a_j, a_k  \in A} d_A(a_i, a_k)  \leqslant d_A(a_i, a_j) + d_A(a_j, a_k)
\label{eq:metric_space_triangle_A}
\end{equation}

\subsubsection{SDR space}
 
\begin{equation}
d_S:S \times S \rightarrow \mathbb{N}
\label{eq:metric_space_distance_metric_S}
\end{equation}

\begin{equation}
\bigwedge_{s_i, s_j \in S} d_S(s_i, s_j)  \geqslant 0
\label{eq:metric_space_postulate_separation_S}
\end{equation}

\begin{equation}
\bigwedge_{s_i, s_j \in S} d_S(s_i, s_j) = 0 \Longleftrightarrow s_i = s_j
\label{eq:metric_space_identity_S}
\end{equation}

\begin{equation}
\bigwedge_{s_i, s_j \in S} d_S(s_i, s_j) = d_S(s_j, s_i)
\label{eq:metric_space_symmetry_S}
\end{equation}

\begin{equation}
\bigwedge_{s_i s_j, s_k  \in S} d_S(s_i, s_k)  \leqslant d_S(s_i, s_j) + d_S(s_j, s_k)
\label{eq:metric_space_triangle_S}
\end{equation}

\subsection{Relationship between Input and SDR spaces}
Input and SDR spaces are correlated (Eq.\ref{eq:Input_and_SDR_relationship}).

\begin{equation}\label{eq:Input_and_SDR_relationship}
\begin{split}
 \bigwedge_{a_i, a_j, a_k, a_w \in A} d_S(f(a_i), f(a_j)) \geqslant d_s(f(a_k), f(a_w) ) \\ \Longleftrightarrow d_A(a_i, a_j) \leqslant d_A(a_k, a_w)
\end{split}
\end{equation}

\subsection{Basic assumptions}
There are several important aspects which must be considered in a process of encoding data:
\begin{itemize}
  \item $ \bigwedge_{a_i,a_j \in A \wedge s_i, s_j \in S: s_i = f(a_i), s_j = f(a_j)}: a_i \approx a_j \Rightarrow 
 s_i \approx s_j $,
  \item $ \bigwedge_{a_i,a_j \in A \wedge s_i, s_j \in S: s_i = f(a_i), s_j = f(a_j)}: a_i = a_j \Rightarrow 
 s_i = s_j $,
  \item $ \bigwedge_{s_i, s_j \in S} n_i = n_j = n$
  \item $ \bigwedge_{s_i, s_j \in S} w_i \approx w_j \approx w$
\end{itemize}

\subsection{Encoder formalism}
The following notation prerequisites are adopted:

\begin{equation}
s \in \{0,1\}^n
\label{eq:sdr_vector_encoder}
\end{equation}

The number of buckets is given by Eq. \ref{eq:number_of_bucket}.
\begin{equation}
buckets = n - w + 1
\label{eq:number_of_bucket}
\end{equation}

The function mapping from the input space to SDR is expressed by Eq. \ref{eq:mapping_function}.
\begin{equation}
f : A \longrightarrow S(n,k) 
\label{eq:mapping_function}
\end{equation}

Input space is limited by the range is given by Eq. \ref{eq:input_space_range}.
\begin{equation}
\bigwedge_{a \in A} a \in \langle val_{min},val_{max}\rangle
\label{eq:input_space_range}
\end{equation}

Formula for a simple numbers encoder is given by Eq. \ref{eq:number_encoder} and \ref{eq:number_encoder_ranges}.
\begin{equation}
\bigwedge_{a \in A} \bigvee_{s \in S , k \in \langle 0, n \rangle}:
k = \lfloor \frac{buckets \cdot (a - val_{min})}{val_{max} - val_{max}}\rfloor
\label{eq:number_encoder}
\end{equation}

\begin{equation}
\left\{\begin{array}{rl}
\bigwedge_{i \in \langle 0, k) \cup (k + w, n \rangle },   & s[i] = 0\\
\bigwedge_{i \in \langle 0, k + w\rangle },   & s[i] = 1\\
\end{array}\right.
\label{eq:number_encoder_ranges}
\end{equation}

\subsection{Spatial Pooler}

This section  will concentrate on mathematical formalism of Spatial Pooler.  The functionality of spatial pooler can be described in a vector and matrix representation, this format of data can improve the efficiency of  the  operations.
The previous section covered encoder formalism definition which is critical since input data for the Spatial Pooler are generated at this stage. The quality of the data directly affects performance of SP.
In article vectors  are defined by lowercase names with an arrow  hat (the transpose of the vector will produce a column vector).  All matrices  will  be  uppercase.  Subscripts  on vectors  and  matrices  are  presented as right bottom indexes to denote internal elements (e.g. $X_{i,j}$ refers to element in $i$ row and $j$ column). Element wise operations are defined by $\odot$ and $\oplus$ operators. 
The I(k) function is indicator function that returns 1 if  event k given as a parameter
is  true  and  0  otherwise.  The input of this function can be matrix or vector, then the output is
matrix or vector, respectively.

The user-defined input parameters are defined in (Table I). These  are  parameters  that  must  be  defined  before  the  initialization of the algorithm. 

\begin{table}
\centering
		\caption{SP symbols.}
		\label{tab:tab2}
	\begin{centering}
		\begin{tabular}{|c|c|}
		\hline
      Symbol    &   Meaning      \\  
     \hline
     n                                                     &   Number of patterns      \\
		\hline
     p                                                     &   Number of inputs (features) in a pattern    \\
		\hline
    m                                                     &   Number of columns   \\
		\hline
    q                                                     &    Number of proximal synapses per column   \\
		\hline

   $\phi_{+}$                                                     &   Permanence increment amount      \\
		\hline

  $\phi_{-}$                                                     &   Permanence decrement amount      \\
		\hline
  $\phi_{\sigma}$                                                     &   Window of permanence initialization      \\
		\hline
  $\rho_{s}$                                                     &   Proximal synapse activation treshold      \\
		\hline
  $\rho_{d}$                                                     &     Proximal dendrite segment activation treshold    \\
		\hline
 $\rho_{c}$                                                     &     Desired column activity level    \\
		\hline
 $s_{duty}$                                                     &     Minimum activity level scaling factor    \\
		\hline
 $s_{boost}$                                                     &     Permanence boosting scaling factor   \\
		\hline
 $\beta_{0}$                                                     &     Maximum boost    \\
		\hline
  $\tau$                                                    &    Duty cycle period    \\
		\hline

		\end{tabular}
	\end{centering}
\end{table}

The terms $s$, $r$, $i$, $j$ and $k$ are integer indices used in article. Theirs vaues are bounded as follows: $s \in [0,n), r \in [0,p), i \in [0,m), j \in [0,m), k \in [0,q)$ .\\

\subsection{Initialization}

Competitive  learning  networks  have typically each  node fully  connected  to  each  input.  The other architectures and techniques like self organizing maps, stochastic gradient networks have single input connected to single node. In Spatial Pooler the  inputs connecting  to  a  particular  column  are  determined  randomly. The density of inputs visible by Spatial Pooler can be computed by using input parameters which defines SP architecture. These rules and dependencies formulas will be described in this section. 
Let  $\vec{c} \in Z^{1 \times m}$ be the vector of columns indices. The $\vec{c}_{i}$ where $i \in [0,m)$
is the column's index at $i$. Let  $I \in {\{0,1\}}^{n \times p}$ be the set of input patterns, such that $I_{s,r}$ indicates $s$ index of $r$ pattern.  \\
Let column synapse index  $CSI \in r^{m \times q}$ be  the  source  column  indices  for  each
proximal synapse on each column, such that $CSI_{i,k}$ is the source column's $\vec{c}_{i}$ index of proximal synapse at index $k$ ($CSI_{i,k}$ refers to specific index in $I_{s}$). \\
For hardware implementation purpose and its memory limitations the modifications to formal description were done. The following changes parameters were defined:
\begin{itemize}
\item $ccon_{i}$ - central index of input to which column $i$ is assign
\item radius - spectrum from $ccon_{i}$ in which the indexes of synapses' inputs are randomly selected
\item $CSI_{i,k} \in {\{ccon_{i} \pm radius\}}^{m\times q}$
\end{itemize}

Given $q$ and $p$,  the  probability  of  a  single  input $I_{s,r}$, connecting to a column is calculated by using \ref{eq:prob}. In \ref{eq:prob}, the probability of an input not connecting is first determined.
That probability is independent for each input. The total probability  of  a  connection  not  being  formed  is  simply  the product of those probabilities. The probability of a connection
forming  is  therefore  the  complement  of  the  probability  of a connection not forming.\\

\begin{equation}
P(i_{r}c_{i})=1-\prod_{k=0}^{q}(1-\frac{1}{p-k})=\frac{q+1}{p}
\label{eq:prob}
\end{equation}

The number of columns that are connected to single input (\ref{eq:n}):
\begin{equation}
ncol_{r}=1-\sum_{i=0}^{m-1}\sum_{k=0}^{q-1}I(CSI_{i,k}==r)
\label{eq:n}
\end{equation}

Expected number of columns to which given input is connected (\ref{eq:enc}): 
\begin{equation}
E[ncol_{r}]=m*P(i_{r}c_{i})
\label{eq:enc}
\end{equation}

The version of above equations for hardware implementation are as follows (equations \ref{eq:probh}, \ref{eq:nch} and \ref{eq:nchol}): \\

\begin{equation}
\begin{split}
P(i_{r}c_{i})=\begin{cases}  1-\prod_{k=0}^{q}(1-\frac{1}{2*radius-k}) & \text{if}  r \in \\ [ccon_{i}-radius, ccon_{i}+radius]\\0  & \text{if}  r \in \\ [0, ccon_{i}-radius] \cup [ccon_{i}+radius, p] \end{cases}
\label{eq:probh}
\end{split}
\end{equation}

\begin{equation}
ncol_{r}=1-\sum_{i=0}^{m-1}\sum_{k=0}^{q-1}I(CSI_{i,k}==r)  
\label{eq:nch}
\end{equation}

\begin{equation}
E[ncol_{r}]=m*P(i_{r}c_{i})
\label{eq:nchol}
\end{equation}

It  is  also possible  to  calculate  the  probability  of  an input never connecting \ref{eq:nvc}.

\begin{equation}
   P(ncol_{r}=0)=(1-P(i_{r}c_{i})^{m}
   \label{eq:nvc}
\end{equation}

The probabilities of connecting or not connecting input to different columns are  independent, it  reduces  to  the  product of  the probability of a input not connecting to a column, taken over all  columns (\ref{eq:ncc} and \ref{eq:encc}).

\begin{equation}
  nccol = \sum_{r=0}^{p-1}I(ncol_{r}==0)
  \label{eq:ncc}
\end{equation}
\begin{equation}
  E[nccol]=p*P(ncol_{r}==0)
  \label{eq:encc}
\end{equation}

Using  equation  \ref{eq:prob} and equation  \ref{eq:nvc},  it  is  possible  to  obtain a  lower  bound  for input parameters $m$ and $q$, by  choosing  those  parameters such  way that  a  certain  percentage  of  input  visibility  is  obtained.  To guarantee observance of all inputs, equation \ref{eq:ncc} must be zero. The desired number of times an input is observed may be determined by using equation {eq:enc}.

 After connecting columns to input,  the  permanences of synapses
 must  be  initialized.  Permanences  were defined to be initialized with a random value close to $\rho_{s}$.  Permanences  should be  randomly  initialized,  with
approximately  half  of  the  permanences  creating  connected
proximal  synapses  and  the  remaining  permanences  creating
potential  (unconnected)  proximal  synapses.  

Let $\phi \in R^{m \times q}$ be defined as the set of permanences for each column, $\phi_{i}$ is set of permanences for proximal synapses for column $\vec{c}_{i}$. $\phi_{i,k}$ is initialized by formula \ref{eq:phi}. Expected permanence value is equal to $\rho_{s}$. Therefore $\frac{q}{2}$  synapses will be connected. The parameter $\rho_{d}$ should be less to give chance each column to be activated at the beginning of learning process.

\begin{equation}
  \phi_{i,k} \sim uniform(\rho_{s}-\phi_{\delta}, \rho_{s}+\phi_{\delta})
\label{eq:phi}
\end{equation}

As initial parameters are set the activation process can be described by mathematical formulas. Let $X \in {\{0,1\}}^{m \times q}$ is the set of inputs for each column, $X_{i}$ set of inputs for column $\vec{c}_{i}$. Let $ac_{i}=\sum_{k}^{q-1}X_{i,k}$ be the random variable of number of active inputs on column $i$. The average number of active inputs on a column is defined by: 
$ac=\frac{1}{m}\sum_{i=0}^{m-1}\sum_{k=0}^{q-1}X_{i,k}$. The $P(X_{i,k})$ is
defined as the probability of the input connected to column $i$ via proximal synapses.  Therefore expected number of active proximal synapses can be computed as follows \ref{eq:eac}:

\begin{equation}
  E[\vec{ac}_{i}]=q*P(X_{i})
\label{eq:eac}
\end{equation}

Let $ActCon_{i,k}=X_{i,k} \cap I(\phi_{i,k} \ge \rho_{s})$ defines the event that proximal synapse $k$ is active and connected to column $i$. Random variable of number of active and connected synapses for column $i$ is define by \ref{eq:acon}:

\begin{equation}
  \vec{actcon}_{i}=\sum_{k=0}^{q-1}ActCon_{i,k}
  \label{eq:acon}
\end{equation}

The probability that synapse is active and connected: $P(ActCon_{i})=P(X_{i,k})*\frac{1}{2}$. Expected number of active and connected synapses for single column is defined as \ref{eq:eacon}:

\begin{equation}
  E[\vec{actcon}_{i}]=q*P(ActCon_{i})
  \label{eq:eacon}
\end{equation}

$Bin(k,n,p)$ is the probability mass function of a binomial distribution ($k$ number of successes, $n$ number of trials, $p$ success probability in each trial). Number of columns with more active inputs than threshold \ref{eq:actc}:

\begin{equation}
  acts=\sum_{i=0}^{m-1}I(\sum_{k=0}^{q-1} X_{i,k} > \rho_{d})
  \label{eq:actc}
\end{equation}

Number of columns with more active and connected proximal synapses than threshold \ref{eq:actcol}: 

\begin{equation}
  actcol=\sum_{i=0}^{m-1}I(\sum_{k=0}^{q-1} ActCon_{i,k} > \rho_{d})
  \label{eq:actcol}
\end{equation}

Let $\pi_{x}$ be the mean of P(x) and $\pi_{ac}$ the mean of $P(ActCon)$ then by \ref{eq:pactcol} and \ref{eq:eactc},  the  summation  computes  the  probability  of  having less  than $\rho_{d}$ active  connected  and active proximal  synapses,  where  the
individual probabilities within the summation follow the PMF
of  a  binomial  distribution.  To  obtain  the  desired  probability,
the  complement  of  that  probability  is  taken.\\
\\
\begin{equation}
  E[acts]=m*[1-\sum_{t=0}^{\rho_{d}-1}Bin(t,q,\pi_{x})]
  \label{eq:eactc}
\end{equation}

\begin{equation}
  P[ActCol]=[1-\sum_{t=0}^{\rho_{d}-1}Bin(t,q,\pi_{ac})]
 \label{eq:pactcol}
\end{equation}

\begin{equation}
  E[actcol]=m*[1-\sum_{t=0}^{\rho_{d}-1}Bin(t,q,\pi_{ac})]
 \label{eq:eactcol}
\end{equation}

\subsection{Overlap}

Let $ConSyn \in {\{0,1\}}^{m \times q}$ and $ConSyn_{i} \in {\{0,1\}}^{1 \times q}$ be the bit mask for proximal synapses connectivity where $ConSyn_{i,k}$ (equation \ref{eq:consyn}).  \\
\\
\begin{equation}
ConSyn_{i,k} =  \begin{cases} 1 & \text{if } \phi_{i,k} \ge \rho_{s}\\ 0 & \text{otherwise}\end{cases}
\label{eq:consyn}
\end{equation}
\\
Let $\vec{b} \in R^{1 \times m}$ be the vector of boost values for each column (${b}_{i}$ is the boost value for $ith$ column). The equation \ref{eq:ovlp} computes $ovlp_{i}$, which is the number of synapses in connected state whose input is activated (line 4 in \ref{alg:1}):\\  
\\
\begin{equation}
\label{eq:ovlp}
\vec{ovlp_{i}}=\vec{X_{i}} \times \vec{ConSyn_{i}}
\end{equation}
\\
Then real overlap $\vec{r}\_ovlp$ value based on boost factor can be computed for each column (equation \ref{eq:rovlp}). The value $r\_ovlp_{i}$ is greater then zero if $ovlp_{i} \ge 0$ is (line 6-9 in \ref{alg:1}).\\
\\
\begin{equation}
\vec{r\_ovlp_{i}} =  \begin{cases} \vec{ovlp_{i}} \cdot b_{i} & \text{if } ovlp_{i} \ge \rho_{d}\\ 0 & \text{otherwise} \end{cases}
\label{eq:rovlp}
\end{equation}
\\
\subsection{Inhibition}

After computing overlap values for each column, the process of activating them is based on set of neighbours of the specified column (defined by inhibition radius). Therefore neighborhood mask matrix is performed as $N \in {\{0,1\}}^{m \times m}$, where $N_{i}$ is the neighborhood of the $ith$ column. Each element in a matrix is indicator of event that column belongs to  neighborhood or not (1 or 0 value). In case of further optimizations the matrix can be reduced to 
$N \in {\{0,1\}}^{m \times inhibition\_radius}$. Let $kmax(S,k)$ be the k-th largest value from set S and $max(\vec{v})$ be the largest value from vector $\vec{v}$. The set of active columns can be represented by vector $\vec{r\_actCol} \in {\{0,1\}}^{1 \times m}$. The vector can be obtained by computing following formula (line 2 in \ref{alg:2}, equation \ref{eq:gam}):\\
\\
\begin{equation}
\vec{\gamma}=max*(kmax(N_{i} \odot r\_ovlp, \rho_{c})) \forall i
\label{eq:gam} 
\end{equation}
\\
then (line 3-7 in \ref{alg:2}, equation \ref{eq:ractcol}):\\
\\
\begin{equation}
\vec{r\_actCol}=I(r\_ovlp_{i} \ge \gamma_{i}) \forall i
\label{eq:ractcol}
\end{equation}

\subsection{Learning}

Learning phase consists of updating permanence values, inhibition radius and boosting factors updating and duty cycle parameters coputing. The permanence values of synapses are updated only when column is activated. Therefore update of synapse can be defined as element wise multiplication of transposed vector of column activations and matrix of values of inputs connected to columns synapses. If inputs are active then permanences are increased by value $\theta_{+}$ otherwise decreased by $\theta_{-}$ (line 6 and 8 in \ref{alg:3}, equations \ref{eq:dphi} and \ref{eq:dphik}).\\
\\
\begin{equation}
\delta \phi=r\_actCol^{T} \odot(\theta_{+}X-(\theta_{\_}\neg X))
\label{eq:dphi}
\end{equation}
\\
\begin{equation}
\delta \phi_{i,k}=r\_actCol_{i}^{T} \odot(\theta_{+}X_{CSI_{i,k}}-(\theta_{\_}\neg X_{CSI_{i,k}}))
\label{eq:dphik}
\end{equation}
\\
\\
The permanence values must been in [0,1] range. The following equation is a rule of updating final permanence values (line 6 and 8 in \ref{alg:3}, equation \ref{eq:phiu}):\\
\\
\begin{equation}
\phi=clip(\phi \oplus \delta \phi, 0,1)
\label{eq:phiu}
\end{equation}
\\
The clip function clips the permanence values in [0,1] range (\ref{eq:clip}). \\
\\
\begin{equation}
clip(k, l, u) = \begin{cases} u & \text{if } k \ge u\\ l & \text{if } \le l\\ k & \text{otherwise} \end{cases}
\label{eq:clip}
\end{equation}
\\
Each column in learning phase updates activeDutyCycle parameter - $\vec{\mu}^{a}_{i}$. The set of these paramters is represented by vector $\vec{\mu}^{a}$. It is worth noticed that history of activation of the columns activation should be stored in a additional structure to remember and update duty cycle parameter in each cycle - $ActDCHist = {\{0,1\}}^{m \times history}$, (only set number of steps before should be remember, history parameter is sliding window width). The activeDutyCycle is computed as follows (equation \ref{eq:acdc}):\\
\\
\begin{equation}
\vec{\mu}^{a}_{i} = \sum_{k=0}^{\tau} ActDCHist_{i,k}
\label{eq:acdc}
\end{equation}
\\
The procedure of $update\_active\_duty\_cycle$ in each cycle can be done by organizing above matrice as cycle list. In each cycle the whole single column is updated. Then the index to the column which will be update in next cycle is incremented by one. If the index will be greater then matrice width the index is set to 0.  
The minimum active duty cycle $\vec{\mu}^{min}$ is computed for boosting purposes by the following equation (\ref{eq:mdc}):\\
\\
\begin{equation}
\vec{\mu}^{min}=s_{duty}*max(H_{i} \odot \vec{\mu}^{a}) \forall i
\label{eq:mdc}
\end{equation}
\\
The maximal active duty cycle of columns in neighbourhood is scaled by $s_{duty}$ factor. \\
The boost factor computation is base on $\mu^{a}, \mu^{min}$ parameters. The boost function should be used when $\mu^{a} \le \mu^{min}$. It should be monotonically decreasing due to $\mu^{a}$ (equations \ref{eq:boost} and \ref{eq:beta}).\\
\\
\begin{equation}
b=\beta(\mu^{a}, \mu^{min}) \forall i
\label{eq:boost}
\end{equation}
\\
\begin{equation}
\beta(\mu^{a}, \mu^{min})=\begin{cases}  \beta_{0} & \text{for }  \mu^{min}=0\\1  & \text{for }  \mu^{a} > \mu^{min}\\ \text{boost function otherwise} \end{cases}
\label{eq:beta}
\end{equation}
\\
The next parameter $\vec{\mu}^{o}$ is $overlapDutyCycle$. It is computed by the same manner like $activeDutyCycle$. Apart from activation indicators the overlap are used. The similar matrice of overlap history is used - $OvlpDCHist$. The permanence boosting definition is based on comparing $\vec{\mu}^{o}$ < $\vec{\mu}^{min}$. If it is true than all input synapses permanence are increased by constant factor (line 8 in \ref{alg:4}, equation \ref{eq:clipb}).\\
\\
\begin{equation}
\phi=clip(\phi \oplus s_{pboost}*I(\vec{\mu}^{o} < \vec{\mu}^{min}), 0,1)
\label{eq:clipb}
\end{equation}
\\
The original inhibition radius presented by Numenta is based on distances between columns and active connected synapses (equation \ref{eq:dist}). Equation \ref{eq:inh} presents how inhibition is computed (sum of distances divided by sum of connected synapses). The inhibition radius can be constant during learning phase or can be changed. It depends of SP mode. Both modes what will be described later should converge to the stable value of inhibition radius or to value with minimal variance. 
\\
\begin{equation}
D=d(pos(c_{i},0), pos(CSI_{i,k}))\odot ConSyn_{i} \forall i \forall k
\label{eq:dist}
\end{equation}
\begin{equation}
inhibition_{i}=max(1,\lfloor \frac{\sum_{i}^{m}\sum_{k}^{q}D_{i,k}}{max(1,\sum_{i}^{m}\sum_{k}^{q}ConSyn_{i,k})} \rfloor)
\label{eq:inh}
\end{equation}

It should be noticed that the spectrum of inhbition radius in case of hardware implementation can be shifted or decreased in some situations. In GPU when columns are processed by thread blocks, boundary threads compare theirs $overlap$ and $activeDutyCycle$ in spectrum of reduced inhibition radius to avoid device memory synchronization \cite{Pietron}.  
During initialization process the mean distance and inhibition is defined by equations \ref{eq:mdist} and \ref{eq:inh}:

\begin{multline}\label{eq:mdist}
mean\_dist(c_{i})= \\ \frac{\frac{end-pos}{input\_size}*q*av_{left} + \frac{pos-start}{input\_size}*q*av_{right}}{q}
\end{multline}

\begin{equation}
inhibition=\frac{\sum mean_{i}}{\frac{1}{2}*q}
\label{eq:inh}
\end{equation}

The initial probability of column activation based on inhibition radius is defined by equation \ref{eq:racol}.

\begin{equation}
P(r\_ActCol)=\frac{k}{inhibition}*P(ActCol)
\label{eq:racol}
\end{equation}

The probability of boosting at initial stage can be computed by equation \ref{eq:pboost}.
\begin{equation}
P(boost)=\frac{inhibition-k}{inhibition}*P(ActCol)
\label{eq:pboost}
\end{equation}

\subsection{Quality of SP}

The Spatial Pooler output representation is in sparse format so number of active columns is $k<<n$ (equations \ref{eq:spc} and \ref{eq:maxsp}). \\
\\
\begin{equation}
|\sum I(ActCol == 1)| = k
\label{eq:spc}
\end{equation}
\begin{equation}
max(k) \simeq 4-5\% (minoverlap == 1 \text{ if sparse input})
\label{eq:maxsp}
\end{equation}
\\
As  SP pattern is learnt we can estimate what input give the same output and its probability. The probability is equals to product of probabilities that for active columns for given pattern the overlap is greater or equal to $minoverlap$ (equation \ref{eq:sim}).\\
\\
\begin{equation}
\prod_{i=0}^{N-k}P(ovlp_{i}<minovlp)*\prod_{i=0}^{k}P(ovlp_{i} \ge minovlp)
\label{eq:sim}
\end{equation}
\\
The single probability can be computed by following equation (\ref{eq:sprob}):\\
\begin{equation}
P(ovlp_{i} \ge minovlp)=\prod_{k=0}^{q} P(X_{CSI_{i,k}->(\phi_{i,k} \ge \rho{s})}==1)
\label{eq:sprob}
\end{equation}

The number of unique patterns that can be represented on $n$ bits with $w$ bits on is defined by (equation \ref{eq:choose}):\\
\\
\begin{equation}
{n \choose{w}} = \frac{p!}{w!(p-w)!}
\label{eq:choose}
\end{equation}
\\
Then we can define the number of codings that can give similar output as learnt pattern by SP (equation \ref{eq:probact} is a number of input codings for actie columns and \ref{eq:probnonact} is number of input codings for non active columns).\\
\\
\begin{equation}
\prod_{i=0}^{act}(2^{(q-\sum I(\phi_{i,k} \ge \rho{s}))}*  {\sum I(\phi_{i,k} \ge \rho{s}) \choose minoverlap}
\label{eq:probact}
\end{equation}

\begin{equation}
\prod_{i=0}^{N-act}(2^{(q-\sum I(\phi_{i,k} \ge \rho{s}))}* \sum_{g=0}^{minoverlap-1} {\sum I(\phi_{i,k} \ge \rho{s}) \choose g-1}
\label{eq:probnonact}
\end{equation}
\\
The $2^{(q-\sum I(\phi_{i,k} \ge \rho{s}))}$ is the number of codings on input to synapses that are learnt zero bit pattern ($\phi_{i,k} < \rho{s}$). The $\sum I(\phi_{i,k} \ge \rho{s}) \choose minoverlap$ is number of codings that input has more bits on than minoverlap on synapses learnt for receiving bits with value 1 ($\phi_{i,k} \ge \rho{s}$).

\subsection{Convergence of SP}
In this section the convergence of SP learning process will be described.  We divided the process 
of learning SP to two different modes. The first one consists of learning each pattern seperately. In this case for each column $c_{i}$ the final state of SP after learning process should be as follows (for $t \rightarrow \infty$):\\
\begin{itemize}
\item $\sum_{0}^{q}X_{i,k} > minovlp$\\
$\phi_{i,k}\rightarrow 1.0$ for $X_{i,k} == 1$\\
$\phi_{i,k}\rightarrow 0.0$ for $X_{i,k} == 0$\\
\item $\sum_{0}^{q}X_{i,k} < minovlp$$\text{,  }$ $\phi_{i,k}\rightarrow 1.0$\\
\end{itemize}

There are three possible starting states at the beginning of learning, the possible transisions from state to other state are as follows (indicated by $ \rightarrow $):
\begin{itemize}
\item not overlap $ \rightarrow $ ($t \rightarrow \infty$) permanence boosting $ \rightarrow $ (overlap $\ge$ minoverlap) if $\sum_{k=0}^{q} X_{i,k} \ge \rho_{s}$ 
\item overlap $\ge$ minOverlap \& no activation $\rightarrow$ overlap boosting (activeDutyCycle value) $\rightarrow$ activation $\rightarrow$ permanence updating
\item overlap $\ge$ minOverlap \& activation $\rightarrow$  permanence updating 
\end{itemize}

It can be noticed that if there are more columns than $k$ parameter with overlap greater or equal than $minoverlap$ value in spectrum of constant inhibition radius than columns are in priority queue (priority is activeDutyCycle) in which they are will be activated in cyclic way (because of overlap boosting). \\
Process of single pattern learning can be run further for next pattern. Before this process learnt columns (columns activated by learnt pattern) should be blocked from permanence boosting (avoid boosting of learnt synapses).  The columns activated (learnt) by previous pattern can ba activated by new pattern only when overlap between inputs of patterns to this column is greater or equal $minoverlap$. Overlap function is defined as follows (equation \ref{eq:overlap}):\\

\begin{equation}
overlap(x,y) = x \times y
\label{eq:overlap}
\end{equation}
where: x and y are binary vectors.

In case of SP learning process of multiple patterns simultaneously there can exist some other situation which can speed up or slow down process of convergence. These all situation are mentioned below:
\begin{itemize}
\item detraction of 1 on single synapse when multiple patterns activate the same column with opposite input value on single synapse $(((r\_ovlp_{i,s}<minoverlap) \& (X_{i,k} == 1)) || (ovlpDC_{i,s}<minActDC_{i,s}))$  $\&$ $((r\_ovlp_{i,s+1}<minoverlap) \& (X_{i,k} == 0))$
\item P(detraction of 0)=(P(Act=1)*P(synapse=1) + P(boost))
\item attraction of 1 on single synapse when multiple patterns activate the same column with the same input value on single synapse $(((r\_ovlp_{i,s}<minoverlap) \& (X_{i,k} == 1)) || (ovlpDC_{i,s}<minActDC_{i,s}))$  $\&$ $((r\_ovlp_{i,s+1} > minoverlap) \& (X_{i,k} == 1))$ 
\item attraction of 0 on single synapse when multiple patterns activate the same column with the same input value on single synapse $((r\_ovlp_{i,s}<minoverlap) \& (X_{i,k} == 0))$  $\&$ $((r\_ovlp_{i,s+1} > minoverlap) \& (X_{i,k} == 0))$
\end{itemize}

There are three possible situations during learning process (multiple pattern learning with constant inhibition radius):
\begin{itemize}
 \item permanence boosting of inputs of columns activated by different patterns, harmful effect but if duty cycle big enough (almost bigger than number of patterns), inputs will be learned 
 \item attraction (equations above) - speeding up learning
 \item detraction (equations above) - slowing down learning
\end{itemize}

In both presented situations (single and multi pattern learning) constant inhibition radius is used. According to the original Numenta algorithm the fluctuations of inhibition radius should decrease during learning process \cite{Mnatzaganian}, but there is possibility that inhibition radius never converge to constant value. In our approach the constant inhibition radius allows to show convergence of learning process. This situation can be achieved by stopping the inhibition radius changing after some learning steps or to change algorithm by the one with radius convergence to stable value. It should be noticed that values of inhibition radius and $k$ should guarantee sparse output at the end of learning.      

\section{Conclusions and future work}
\label{section:conclusions}

The presented HTM model is a new architecture in deep learning domain inspired by human brain. Initial results show \cite{Cui} that it can be at least efficient like other machine and deep learning models. Additionally, our earlier research \cite{Pietron} showed that it can be significantly speed up by hardware accelerators.
The presented formalism is one of the first article with full mathematical description of HTM Spatial Pooler. The formal description will help to parameterized the model. According to given encoder and its input distribution characteristic it is possible by formal model to estimate number of learning cycles, probability of patterns attraction, detraction etc. 
Further work the should concentrate on extending the formalism by accurate proofs of convergence of learning process. Then formal description of temporal pooler should be added.

\end{document}